# Accelerated Evaluation of Automated Vehicles Safety in Lane Change Scenarios based on Importance Sampling Techniques

Ding Zhao, Henry Lam, Huei Peng, Shan Bao, David J. LeBlanc, Kazutoshi Nobukawa, and Christopher S. Pan

*Abstract*—Automated vehicles (AVs) must be evaluated thoroughly before their release and deployment. A widely-used evaluation approach is the Naturalistic-Field Operational Test (N-FOT), which tests prototype vehicles directly on the public roads. Due to the low exposure to safety-critical scenarios, N-FOTs are time-consuming and expensive to conduct. In this paper, we propose an accelerated evaluation approach for AVs. The results can be used to generate motions of the primary other vehicles to accelerate the verification of AVs in simulations and controlled experiments. Frontal collision due to unsafe cut-ins is the target crash type of this paper. Human-controlled vehicles making unsafe lane changes are modeled as the primary disturbance to AVs based on data collected by the University of Michigan Safety Pilot Model Deployment Program. The cut-in scenarios are generated based on skewed statistics of collected human driver behaviors, which generate risky testing scenarios while preserving the statistical information so that the safety benefits of AVs in non-accelerated cases can be accurately estimated. The Cross Entropy method is used to recursively search for the optimal skewing parameters. The frequencies of occurrence of conflicts, crashes and injuries are estimated for a modeled automated vehicle, and the achieved accelerated rate is around 2,000 to 20,000. In other words, in the accelerated simulations, driving for 1,000 miles will expose the AV with challenging scenarios that will take about 2 to 20 million miles of real-world driving to encounter. This technique thus has the potential to reduce greatly the development and validation time for AVs.

*Index Terms*—Automated vehicles, importance sampling, crash avoidance, active safety systems, lane change, AEB

## I. INTRODUCTION

Automated vehicle (AV) technologies are actively studied by many companies because of their potential to save fuel, reduce crashes, ease traffic congestion, and provide better mobility, especially to those who cannot drive [1]. Currently, almost all major automakers have research and development programs on AVs. By 2030, it is estimated that the sales of AVs may reach $87 billion dollars [2].

National Highway Traffic Safety Administration defines five levels of AV automation [3]. AVs are quickly being developed from level 0 automation, which conducts no driving tasks and up, possibly all the way to level 4 automation, which monitors the driving environment performs all dynamic driving duties. As the level of automation goes up, AVs need to deal with many uncertainties/disturbances in the real world, including imperfect human driver behaviors. AVs are projected to penetrate the market gradually and will co-exist with human-controlled vehicles (HVs) for decades [4]. During this transitional period, AVs will interact primarily with HVs. It is estimated that 70-90% of motor vehicle crashes are due to human errors [5], [6]. However, AVs can have their own crash modes. A practical and effective evaluation of the safety performance of AVs should consider their interactions with HVs.

Manuscript received on XXXXXX. This work was supported by the U.S. National Institute for Occupational Safety Health (NIOSH) under Grant F031433.

Ding Zhao and Huei Peng are with the Mechanical Engineering Department, University of Michigan, Ann Arbor, MI 48109 USA (e-mail: zhaoding@umich.edu and hpeng@umich.edu).

Henry Lam is with the Department of Industrial and Operations Engineering, University of Michigan, Ann Arbor, MI 48109 USA (e-mail: khlam@umich.edu).

Shan Bao, David J. LeBlanc and Kazutoshi Nobukawa are with the University of Michigan Transportation Research Institute, Ann Arbor, MI 481099 USA (e-mail: shanbao@umich.edu, leblanc@umich.edu and knobukaw@umich.edu).

Christopher S. Pan is with the Division of Safety Research, NIOSH, CDC Morgantown, WV 26505 USA (e-mail: syp4@cdc.gov).

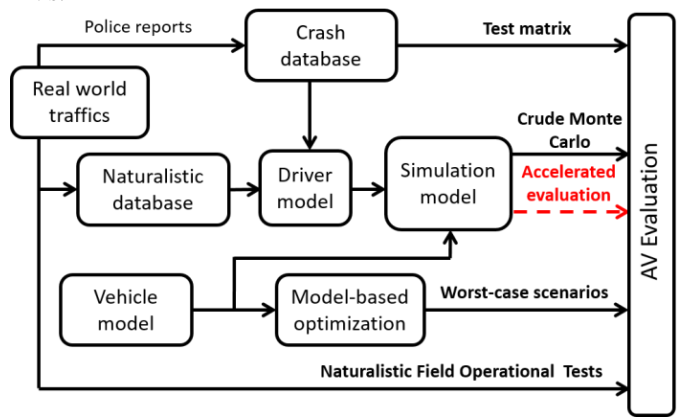

Fig. 1. Summary of evaluation approaches for automated vehicles

Approaches for AV evaluation can be summarized into four categories as shown in Fig. 1. One approach to studying the interactions between AVs and HVs is through Naturalistic Field Operational Tests (N-FOT) [7]. In an N-FOT, data is collected from a number of equipped vehicles driven under naturalistic conditions over an extended period of time [8]. Several N-FOT projects [9]–[16] have been conducted in the U.S. and Europe. Conducting an N-FOT to evaluate an AV function typically involves non-intrusive conditions, i.e., the test drivers were told to drive as they normally do on public roads. This test approach suffers from several limitations. An obvious problem is the time needed. Under naturalistic conditions, the level of exposure to dangerous events is very low. In the U.S., there were 5.7 million police-reported motor vehicle crashes and 30,057 fatal crashes in 2013, while the vehicles traveled a total of 2.99 trillion miles [17]. This translates to approximately 0.53 million miles for a police-reported crash and 99 million miles for a fatal crash. Since the average mileage driven annually by licensed drivers is 14,012 miles [17], one needs to drive on average 38 years to experience a police-reported crash and 6,877 years for a fatal crash. Because of this low exposure rate, the N-FOT projects need a large number of vehicles, long test duration, and a large budget. According to Akamatsu et al. [18], an N-FOT "cannot be conducted with less than $10,000,000". A more efficient approach for AV evaluation is needed.

Some researchers built stochastic models based on the big data obtained from N-FOTs and ran Monte Carlo simulations to evaluate AVs. Yang *et al.* [19] and Lee [20] evaluated collision avoidance systems by replaying segments extracted from the Road-Departure Crash-Warning (RDCW) FOT and Intelligent Cruise Control (ICC) FOT naturalistic driving databases. Woodrooffe *et al.* [21] generated 1.5 million forward collision scenarios based on naturalistic driving conflicts and used them to evaluate collision warning and collision mitigation braking technologies on heavy trucks. Reusing the N-FOT data in simulations can avoid the large budget for N-FOTs. However, even for computer simulations, low exposure to safety critical scenarios is still an issue.

The test matrix approach has been the basis of many test procedures, such as the AEB (Autonomous Emergency Braking) test protocol [22] of the Euro New Car Assessment Program (Euro-NCAP). Much development work was done to advance this evaluation approach including CAMP [23], HASTE [24], AIDE [25], TRACE [26], APROSYS [27] and ASSESS [28]. The test scenarios are frequently selected based on national crash databases [29], such as GES (General Estimates System) [30], NMVCCS (National Motor Vehicle Crash Causation Survey) [31] and EDR (Event Data Recorder databases) [32]. A systematic review of this approach can be found in [8]. The main benefits of this test method are that it is repeatable, reliable, and can be finished in a reasonable amount of time. However, it is not clear how the selected test scenarios correlate with real-world conditions, especially when human interaction is involved [8], [33]. Moreover, because all test scenarios are fixed and predefined, AVs can be tuned to achieve good performance in these tests, but their behaviors under broader conditions are not adequately assessed [34].

Another approach, the Worst-Case Scenario Evaluation (WCSE) methodology, has been studied by Ma *et al.* [35], Ungoren *et al.* [36] and Kou [37] to identify the most challenging scenarios using model-based optimization techniques. While the worst-case evaluation method can identify the weakness of a vehicle control system, it does not consider the probability of occurrence of the worst-case scenarios. Therefore, the worst case evaluation results do not provide sufficient information about the risk in the real world and may not be the fairest way to compare different designs.

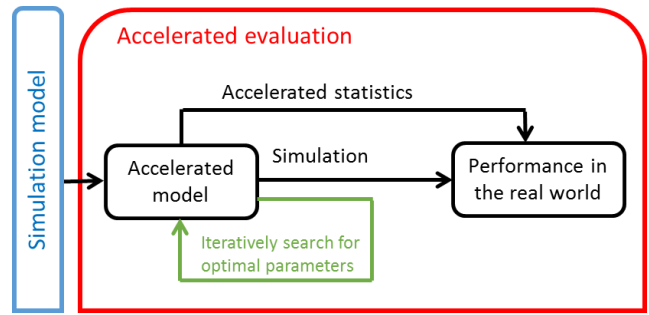

Fig. 2. Procedure of the accelerated evaluation method

In a previous work [38], we proposed the accelerated evaluation concept and applied it to the car-following scenarios. The crash rate in the real world was estimated based on the national crash database. In [39], we introduced the Importance Sampling techniques to improve the reliability and accuracy of the estimation, in which the parameters in the accelerated tests were tuned by hands. In this paper, we further proposed an automated method to search for the best way to morph the original lane change behavior statistics. As shown in Fig. 2, first, HVs are modeled based on data extracted from N-FOT databases to represent the human driving behaviors. Second, an accelerated model is constructed by modifying the probability density functions of the stochastic variables to promote riskier lane change behaviors. Third, the optimal parameters of the accelerated model are obtained through an iterative search. Finally, the 'amplified' results together with the statistics in the accelerated model are used to calculate the performance of the host vehicles in real world driving. The contribution of this paper is that we proposed the Accelerated Evaluation of AV procedure which provides high accuracy and accelerated evaluation using Importance Sampling theory and the Cross Entropy method. To the best of our knowledge, we are the first group to apply these techniques to create test scenarios to evaluate AV safety and calculate social benefits. The meaningfulness of doing this is not only to accelerate the simulation, but also to provide a way to objectively identify critical test scenarios that can be used in other types of evaluation platforms such as driving simulator, on-track tests, or hardware-in-the-loop tests.

## II. LANE CHANGE MODELS BASED ON NATURALISTIC DRIVING

The lane change (cut-in) scenario is used as an example to show the benefits of the proposed accelerated evaluation approach. Lane change, defined as a vehicle moving from one lane to another in the same direction of travel [40], can cause a frontal collision crash for the following vehicle when the time gap is too short. Successful completion of a lane change requires attention to the vehicles in both the original lane and the adjacent lane [41]. In the US, there are between 240,000 and 610,000 reported lane-change crashes, resulting in 60,000 injuries annually [40]. Few protocols have been published regarding the evaluation of AVs (e.g., AEB systems) under lane change scenarios.

Human drivers' lane change behaviors have been analyzed and modeled for more than half a century. Early studies based on controlled experiments usually have short test horizons and limited control settings [42]. More recently, researchers started to use large scale N-FOT databases to model the lane change behaviors. Lee *et al.* [42] examined steering, turn signal and brake pedal usage, eye glance patterns, and safety envelope of 500 lane changes. The 100-Car Naturalistic Driving Study analyzed lane change events leading to rear-end crashes and near-crashes [40]. Zhao *et al.* [43] analyzed the safety critical variables in mandatory and discretionary lane changes for heavy trucks [12]. Most of these studies are based on hundreds of lane changes. We use the data collected in the Safety Pilot Model Deployment (SPMD) project, which contains more than 400,000 lane changes.

### A. Identification of the Lane Change Events

In this research, we developed a lane change statistical model and demonstrated its use for accelerated evaluation of a frontal collision avoidance algorithm. The data used is from the Safety Pilot Model Deployment database [44]. The SPMD program aims to demonstrate connected vehicle technologies in a real-world environment. It recorded naturalistic driving of 2,842 equipped vehicles in Ann Arbor, Michigan for more than two years. As of April 2015, 34.9 million miles were logged, making SPMD one of the largest public N-FOT databases ever.

As shown in Fig. 3, a lane change was detected and recorded by an SPMD vehicle when the Lane Change Vehicle (LCV) crosses the lane markers. In the SPMD program, 98 sedans are equipped with Data Acquisition System and MobilEye® [45], which provides: a) relative position to the lane change vehicle (range), and b) lane tracking measures pertaining to the lane delineation both from the painted boundary lines and road edge characteristics. The error of range measurement is around 10 % at 90 m and 5 % at 45 m [46].

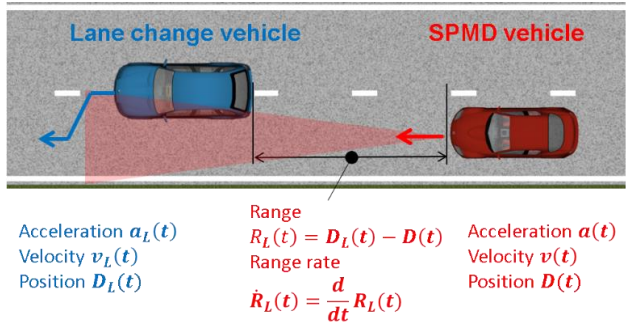

Fig. 3. Lane change scenarios that may cause frontal crashes

To ensure consistency of the used dataset, the following criteria were applied:

- $v(t_{LC}) \in (2\ m/s, 40\ m/s)$
- $v_L(t_{LC}) \in (2\ m/s, 40\ m/s)$     (1)
- $R_L(t_{LC}) \in (0.1\ m, 75\ m)$

where $t_{LC}$ is the time when the center line of the LCV crosses the lane markers; $v_L$ and $v$ are the velocities of the LCV and the SPMD vehicle; $R_L$ is the range defined as the distance between the rear edge of the LCV and the front edge of the SPMD vehicle. 403,581 lane changes were detected in total. Fig. 4 shows the locations of the identified lane changes.

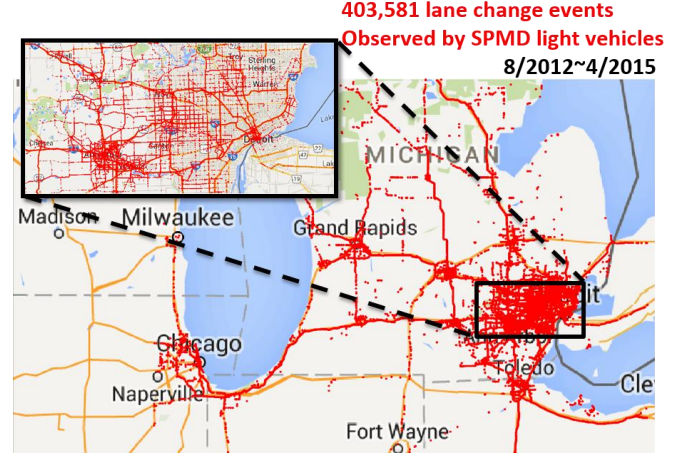

Fig. 4. Recorded lane change events from the SPMD database

### B. Lane Change Models

A lane change can be divided into three phases: the decision to initiate a lane change, gap (range) acceptance, and lane change execution [42]. In this research, we focus on the effects of gap acceptance, which is mainly captured by three variables: $v_L(t_{LC})$, $R_L(t_{LC})$ and Time To Collision (TTC) of AVs, defined as

$$TTC_L = -\frac{R_L}{\dot{R}_L} \quad (2)$$

where $\dot{R}_L$ is the derivative of $R_L$. In the following, unless mentioned specifically, $v_L$, $R_L$ and $TTC_L$ are the variables at $t_{LC}$.

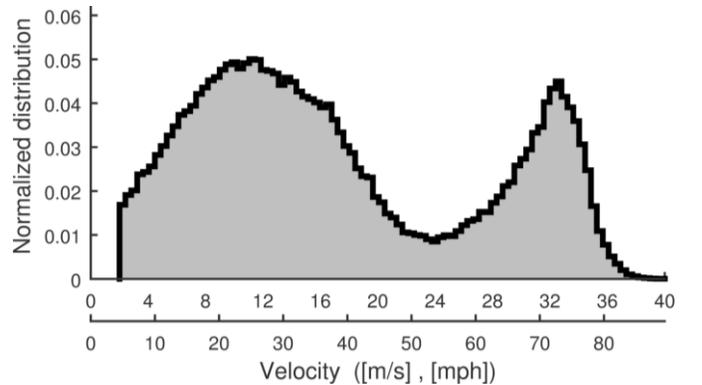

Fig. 5. Distributions of $v_L(t_{LC})$ of lane change events used in our model

The distribution of $v_L$ is shown in Fig. 5. The division of highways and local roads is embodied by the bimodal shape of the histogram. $v_L$ is assumed to remain constant during the lane change. Only the events with a negative range rate are used to build the lane change model. Out of 403,581 lane change events, 173,692 are with negative range rate.

Larger $R_L$ and $TTC_L$ indicate the scenario is safer which are the majority cases in naturalistic driving, while Smaller $R_L$ and $TTC_L$ indicate the scenario is less safe and rarer. Therefore, we define the variables of interest as reciprocal of $R_L$ and $TTC_L$ to put the rare events in the tail of the distribution to naturally fit the naturalistic driving statistics. To capture the influence of vehicle speed on range and TTC, we divided lane change events into low, medium. and high velocity conditions. Fig. 6 shows that $v_L$ has little influence on the distribution of $R_L^{-1}$. We use a standard Matlab package [47] to search for a proper distribution to fit $R_L^{-1}$, which examines 17 different types of distributions and examine goodness-of-fit by using Bayesian Information Criterion [48]. Fig. 7 illustrates the fitting of $R_L^{-1}$ using a Pareto distribution defined as

$$f_{R_L^{-1}}\left(x \middle| k_{R_L^{-1}}, \sigma_{R_L^{-1}}, \theta_{R_L^{-1}}\right) = \frac{1}{\sigma_{R_L^{-1}}}\left(1 + k_{R_L^{-1}} \frac{x - \theta_{R_L^{-1}}}{\sigma_{R_L^{-1}}}\right)^{-1 - 1/k_{R_L^{-1}}} \quad (3)$$

where the shape parameter $k_{R_L^{-1}}$, the scale parameter $\sigma_{R_L^{-1}}$, and the threshold parameter $\theta_{R_L^{-1}}$ are all positive. Note that, due to the physical limitations mentioned in (1), the Pareto distribution in Eq. (3) is in fact truncated at 1/0.1 m$^{-1}$ and 1/75 m$^{-1}$. For the sake of conciseness, we show the untruncated version throughout this paper. The same holds for all other fitted distributions in this paper.

The histograms of $TTC_L^{-1}$ for different velocity intervals are shown in Fig. 8. As the vehicle speed increases, the mean of $TTC_L^{-1}$ decreases. Based on the analysis using Matlab fitting package [47], $TTC_L^{-1}$ can be approximated by both Pareto distribution and exponential distribution with 0.23 % relative difference in BIC. We used the exponential distribution

$$f_{TTC_L^{-1}}\left(x \middle| \lambda_{TTC_L^{-1}}\right) = \frac{1}{\lambda_{TTC_L^{-1}}} e^{-x/\lambda_{TTC_L^{-1}}} \quad (4)$$

for simplicity, where the scaling factor $\lambda_{TTC_L^{-1}}$ varies with the speed of the LCV. Here we define $\lambda_{TTC_L^{-1}}$ as the mean value instead of the rate of the exponential distribution, because mean value has more intuitive physical meaning.

The dependence of $\lambda_{TTC_L^{-1}}$ on vehicle speed is shown in Fig. 9. As the vehicle speed increases, $\lambda_{TTC_L^{-1}}$ decreases. The blue circles represent $\lambda_{TTC_L^{-1}}$ at the center points of $v_L$ intervals. We use linear interpolation and extrapolation to create smooth $\lambda_{TTC_L^{-1}}$ for all vehicle speeds.

The effect of range on TTC is very limited, as can be seen in Fig. 10. This indicates that $R_L$ and $TTC_L$ can be modeled independently given the same $v_L$. $\dot{R}_L$ can then be calculated from Eq. (5).

$$\dot{R}_L = -\frac{TTC_L^{-1}}{R_L^{-1}} \quad (5)$$

Finally, the velocity of the host vehicle $v$ can be calculated from

$$v = v_L - \dot{R}_L \quad (6)$$

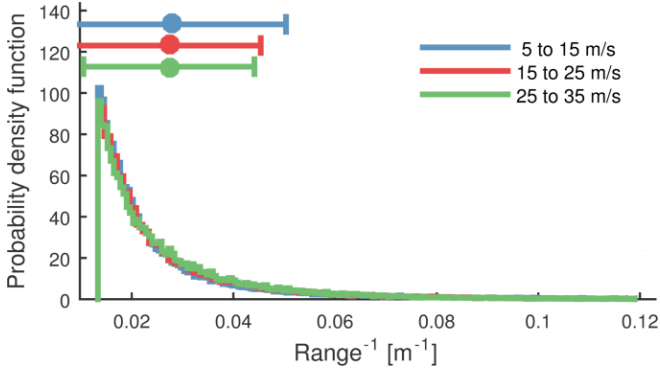

Fig. 6.  Distributions of $R_L^{-1}(t_{LC})$ at different vehicle forward speeds

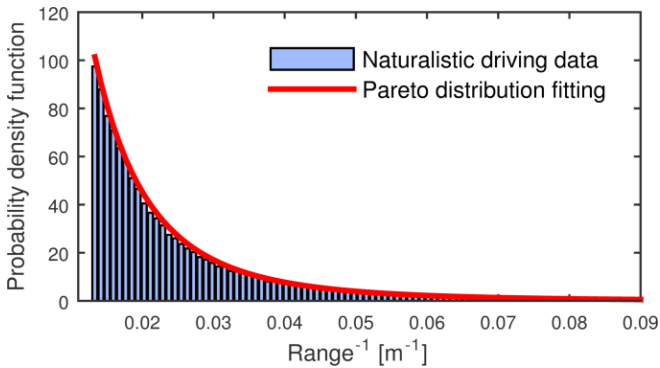

Fig. 7.  Fitting results of $R_L^{-1}(t_{LC})$ using the Pareto distribution

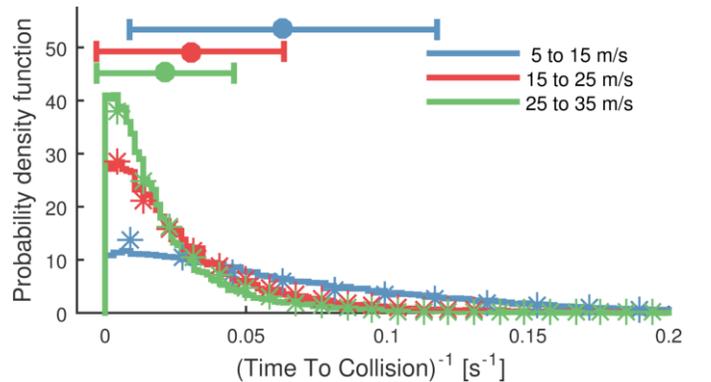

Fig. 8.  Distribution of $TTC_L^{-1}(t_{LC})$ at different lane change vehicle speeds

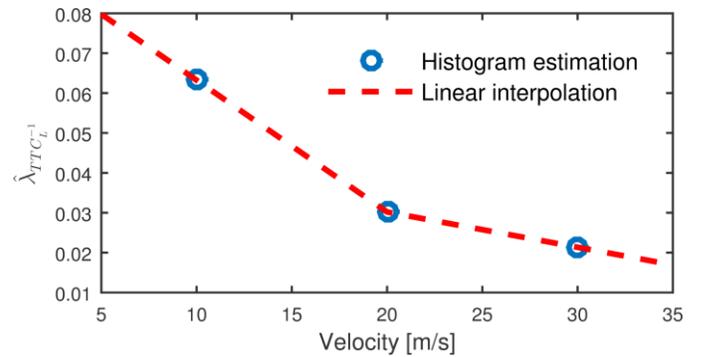

Fig. 9.  Model parameters for $TTC_L(t_{LC})$

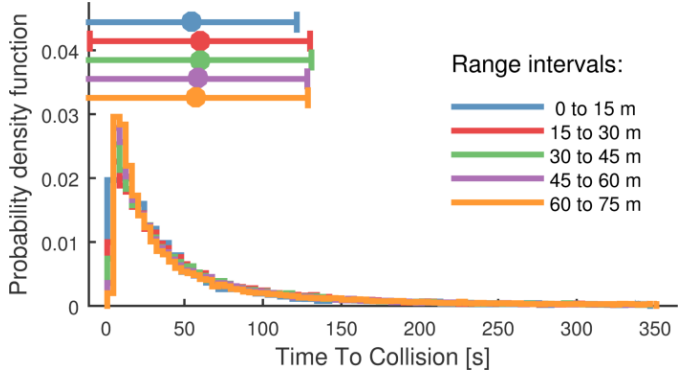

Fig. 10. Distribution of $TTC_L(t_{LC})$ at different range intervals

In summary, the lane change events are generated in the following order: a) generate $v_L$ based on the empirical distributions shown in Fig. 5; b) generate $R_L^{-1}$ using Fig. 7; c) generate $TTC_L^{-1}$ using the Exponential distribution with parameters shown in Fig. 9; and finally d) calculate $v$ using Eqs. (5) and (6).

## III. ACCELERATED EVALUATION

Monte Carlo techniques can be used to simulate driving conditions using a stochastic model, but a naïve implementation will take a long time to execute. The key of accelerated evaluation is to skew the statistics of the Monte Carlo samples but still be able to maintain statistical correctness and accuracy. In this section, we first show the limitation of the 'crude' Monte Carlo (CMC) in simulating events with small probability (rare events). We then introduce the Importance Sampling (IS) concept. Thirdly, we show how to apply IS to evaluate AVs in lane change scenarios. Finally, we introduce the Cross Entropy method to optimize the use of IS.

### A. Monte Carlo Estimation

Monte Carlo method [49] typically aims to generate unbiased statistical samples to estimate the expected value of a stochastic process. Let $\Omega$ be the sample space for all possible events, and $\mathcal{E} \subset \Omega$ be the rare events of interest, e.g., the occurrence of a crash. Let $x$ be a random vector describing the motions of the lane change vehicle. The indicator function of the event $\mathcal{E}$ is defined as

$$I_{\mathcal{E}}(x) = \begin{cases} 1, & \text{if } x \in \mathcal{E} \\ 0, & \text{otherwise} \end{cases} \quad (7)$$

Our goal is to estimate the probability of $\mathcal{E}$ happening, i.e.

$$\gamma = P(\mathcal{E}) = E(I_{\mathcal{E}}(x)) \quad (8)$$

The CMC approach generates independent and identically distributed samples $x_1, x_2, \dots, x_n$ of $x$, and then calculate the sample average

$$\hat{\gamma}_n = \frac{1}{n} \sum_{i=0}^{n} I_{\mathcal{E}}(x_i) \quad (9)$$

We state some statistical properties of CMC. First, under mild conditions, the Strong Law of Large Numbers [49] holds, i.e.

$$P\left(\lim_{n \to \infty} \hat{\gamma}_n = \gamma\right) = 1 \quad (10)$$

Moreover, the Central Limit Theorem [49] implies that, when $n$ is large, $\hat{\gamma}_n$ follows approximately the normal distribution $\mathcal{N}(E(\hat{\gamma}_n), \sigma^2(\hat{\gamma}_n))$ with the mean

$$E(\hat{\gamma}_n) = E\left(\frac{1}{n} \sum_{i=0}^{n} I_{\mathcal{E}}(x_i)\right) = \gamma \quad (11)$$

and variance

$$\sigma^2(\hat{\gamma}_n) = \text{Var}(\hat{\gamma}_n) = \text{Var}\left(\frac{1}{n} \sum_{i=0}^{n} I_{\mathcal{E}}(x_i)\right)$$
$$= \frac{1}{n^2} \sum_{i=0}^{n} \text{Var}\left(I_{\mathcal{E}}(x_i)\right) = \frac{\gamma(1-\gamma)}{n} \quad (12)$$

The accuracy of the estimation is represented by the relative half-width, which is the half-width of the confidence interval relative to the probability to be estimated. With the Confidence Level at $100(1-\alpha)\%$, the relative half-width of $\hat{\gamma}_n$ is defined as

$$l_r = \frac{l_\alpha}{\gamma} \quad (13)$$

where $l_\alpha$ is the half-width given by

$$l_\alpha = z_\alpha \sigma(\hat{\gamma}_n) \quad (14)$$

and $z_\alpha$ is defined as

$$z_\alpha = \Phi^{-1}(1 - \alpha/2) \quad (15)$$

where $\Phi^{-1}$ is the inverse cumulative distribution function of $\mathcal{N}(0,1)$. To ensure $l_r$ is smaller than a constant $\beta$, we need

$$\frac{l_\alpha}{\gamma} = \frac{z_\alpha \sigma(\hat{\gamma}_n)}{\gamma} = \frac{z_\alpha}{\gamma} \sqrt{\frac{\gamma(1-\gamma)}{n}} = z_\alpha \sqrt{\frac{1-\gamma}{\gamma n}} \quad (16)$$
$$\leq \beta$$

In other words,

$$n \geq \frac{z_\alpha^2}{\beta^2} \cdot \frac{1-\gamma}{\gamma} \quad (17)$$

Eq. (17) reveals that when $\mathcal{E}$ is rare, $i.e. \gamma 0$, the required test number $n$ goes to infinity. This means that a huge test number is required to maintain a satisfactory half-width relative to the magnitude of a rare event probability $\gamma$. This is the reason why CMC is slow.

### B. Importance Sampling (IS)

IS is a so-called variance reduction technique that is effective in handling rare events. IS has been successfully applied to evaluate critical events in reliability [50], finance [51], insurance [52], and telecommunication networks [53]. General overviews about IS can be found in [54]–[56].

To explain the concept of IS, we denote $f(x)$ as the original joint density function of the random vector $x$. The core idea of

IS is to replace $f(x)$ with a new density $f^*(x)$ that has a higher likelihood for the rare events to happen. Using a different distribution, however, leads to biased samples, and the key of IS is to provide a mechanism to compensate for this bias and compute correct crash rate at the end.

We describe this mechanism as follows. First, we define the likelihood ratio L (Radon-Nikodym derivative [57]) as

$$L(x) = \frac{f(x)}{f^*(x)} \quad (18)$$

The probability of $\mathcal{E}$ satisfies

$$\begin{aligned} P(\mathcal{E}) &= E_f(I_\mathcal{E}(x)) \\ &= \int I_\mathcal{E}(x) f(x) dx \\ &= \int [I_\mathcal{E}(x) L(x)] f^*(x) dx \\ &= E_{f^*}(I_\mathcal{E}(x) L(x)) \end{aligned} \quad (19)$$

One required condition for Eq. (19) to hold is that $f^*(x)$ must be absolutely continuous with respect to $f(x)$ within $\mathcal{E}$, i.e.

$$\forall x \in \mathcal{E}: f^*(x) = 0 \Rightarrow f(x) = 0 \quad (20)$$

which guarantees the validity of L in Eq. (18). The IS sample is $I_\mathcal{E}(x_i) L(x_i)$ where $x_i$ is generated under $f^*(x)$, which is an unbiased estimator for $\gamma$. The overall IS estimator for test number $n$ is then

$$\hat{\gamma}_n = \frac{1}{n} \sum_{i=0}^{n} I_\mathcal{E}(x_i) L(x_i) \quad (21)$$

Note that although a continuous model is used in this paper, similar approaches can be applied to the discrete model as well.

Now consider the relative half-width of CI constructed by IS

$$\begin{aligned} l_r^* &= \frac{l_\alpha}{\gamma} = \frac{z_\alpha \sigma(\hat{\gamma}_n)}{\gamma} = \frac{z_\alpha \sqrt{E_{f^*}(\hat{\gamma}_n^2) - E_{f^*}^2(\hat{\gamma}_n)}}{\gamma \sqrt{n}} \\ &= \frac{z_\alpha \sqrt{E_{f^*}(I_\mathcal{E}^2(x) L^2(x)) - \gamma^2}}{\gamma \sqrt{n}} \\ &= \frac{z_\alpha}{\sqrt{n}} \sqrt{\frac{E_{f^*}(I_\mathcal{E}^2(x) L^2(x))}{\gamma^2} - 1} \leq \beta \end{aligned} \quad (22)$$

The required minimum test number is then

$$n \geq \frac{z_\alpha^2}{\beta^2} \left( \frac{E_{f^*}(I_\mathcal{E}^2(x) L^2(x))}{\gamma^2} - 1 \right) \quad (23)$$

When $f^*(x)$ is properly chosen, $E_{f^*}(I_\mathcal{E}^2(x) L^2(x))$ can be close to $\gamma^2$, resulting in a smaller number of tests (i.e., the evaluation is accelerated).

*C. Accelerated Evaluation of Automated Vehicles in Lane Change Scenarios*

When a slower lane changing vehicle cut in front of the AV, the events of interest are defined as

$$\mathcal{E} = \{\min(R_L(t)) < R_\mathcal{E} | t_{LC} < t \leq t_{LC} + T_{LC}\} \quad (24)$$

where $T_{LC}$ represents duration of the lane change event; $R_\mathcal{E}$ is the critical range. Eq. (24) means that if the minimum range is smaller than $R_\mathcal{E}$ anytime during the lane change event, this lane change belongs to the $\mathcal{E}$ set.

The random vector $x$ consists of three random variables $[v_L, TTC_L^{-1}, R_L^{-1}]$. $v_L$ is generated using the empirical distributions shown in Fig. 5. The IS approach considers the modified probability density functions of $TTC_L^{-1}$ and $R_L^{-1}$ denoted by $f^*_{TTC_L^{-1}}(x)$ and $f^*_{R_L^{-1}}(x)$. The likelihood ratio is then

$$L(R_L^{-1} = x, TTC_L^{-1} = y) = \frac{f_{R_L^{-1}}(x) f_{TTC_L^{-1}}(y)}{f^*_{R_L^{-1}}(x) f^*_{TTC_L^{-1}}(y)} \quad (25)$$

From Eq. (19), the probability of $\mathcal{E}$ can be estimated as

$$P(\mathcal{E}) = E_f(I_\mathcal{E}(x)) = E_{f^*}(I_\mathcal{E}(x) L(x)) \quad (26)$$

The only task left is then to construct proper $f^*_{TTC_L^{-1}}(x)$ and $f^*_{R_L^{-1}}(x)$ to accelerate the evaluation procedure.

*D. Searching for optimal IS distributions with the Cross Entropy approach*

The choice of IS distribution is critical to the success of the IS method. The Cross Entropy (CE) method, first proposed by Rubinstein [58], is an iterative search procedure to find good IS distribution within a prescribed parametric family.

To understand how CE works, we first point out an important observation: the theoretical optimal IS distribution is always the conditional distribution given that the rare event of interest happens, namely

$$f^*_{zv}(x) = \begin{cases} \frac{f(x)}{\gamma}, & I_\mathcal{E}(x) = 1 \\ 0, & I_\mathcal{E}(x) = 0 \end{cases} \quad (27)$$

With $f^*_{zv}(x)$, any sampled $x$ leads to a rare event so that the indicator function $I_\mathcal{E}(x)$ constantly equals to one. The likelihood ratio

$$L_{zv}(x) = \frac{f(x)}{f^*_{zv}(x)} = \gamma \quad (28)$$

The probability of the rare events is calculated by

$$\hat{\gamma}_n = \frac{1}{n} \sum_{i=0}^{N} I_\mathcal{E}(x_n) L(x_n) = \frac{1}{n} \sum_{i=0}^{n} \gamma = \gamma \quad (29)$$

In other words, $\hat{\gamma}_n$ equals to $\gamma$ for all $n$. The distribution $f^*_{zv}(x)$ is optimal in the sense that any sample generated from it has zero variance, and hence the required test number to construct confidence level to any precision is 1; thus it is also known as the zero variance IS distribution [56]. Unfortunately, this distribution cannot be implemented directly because it requires the knowledge of $\gamma$, which is exactly what we want to estimate. However, it provides a benchmark to get good IS distributions: A good IS distribution should be close to the zero-variance distribution.

To describe how CE works, we define the Kullback–Leibler (KL) divergence

$$f_{KL}\left(\tilde{f}_\vartheta(x), f_{zv}^*(x)\right) = \int \log\left[\frac{f_{zv}^*(x)}{\tilde{f}_\vartheta(x)}\right] f_{zv}^*(x)dx \quad (30)$$

as a measure of the difference between $\tilde{f}_\vartheta(x)$ and $f_{zv}^*(x)$. The idea of CE is to find an IS distribution over the family of distributions $\tilde{f}_\vartheta(x)$ (controlled by $\vartheta$) that has the minimum KL divergence with $f_{zv}^*(x)$, i.e.

$$\vartheta^* = \arg\min_\vartheta f_{KL}\left(\tilde{f}_\vartheta(x), f_{zv}^*(x)\right) \quad (31)$$

Substituting Eq. (30) into Eq. (31), we have

$$\begin{aligned}\vartheta^* &= \arg\min_\vartheta \int \log\left[\frac{f_{zv}^*(x)}{\tilde{f}_\vartheta(x)}\right] f_{zv}^*(x)dx \\ &= \arg\min_\vartheta \int \{\log[f_{zv}^*(x)]\, f_{zv}^*(x) \\ &\quad - \log[\tilde{f}_\vartheta(x)]\, f_{zv}^*(x)\}dx\end{aligned} \quad (32)$$

Note the first term inside the integration is independent of $\vartheta$, Eq. (32) can be simplified to

$$\vartheta^* = \arg\max_\vartheta \int \log[\tilde{f}_\vartheta(x)] f_{zv}^*(x)dx \quad (33)$$

Substituting Eq. (27) into Eq. (33), we have

$$\begin{aligned}\vartheta^* &= \arg\max_\vartheta \int \log[\tilde{f}_\vartheta(x)] \frac{f(x)}{P(\mathcal{E})} I_\mathcal{E}(x)dx \\ &= \arg\max_\vartheta \int \log[\tilde{f}_\vartheta(x)] f(x) I_\mathcal{E}(x)dx\end{aligned} \quad (34)$$

The CE method is an iterative scheme to sequentially improve the IS distribution and optimize $\vartheta^*$ using Eq. (33). At the $i^{th}$ iteration, we use $\tilde{f}_{\vartheta_i}(x)$ as the IS distribution to run the Monte Carlo. Then, letting $\tilde{L}_{\vartheta_i}(x) = f(x)/\tilde{f}_{\vartheta_i}(x)$, from Eq. (34), $\vartheta_{i+1}$ can be derived as

$$\begin{aligned}\vartheta_{i+1} &= \arg\max_\vartheta \int \log[\tilde{f}_\vartheta(x)] \tilde{L}_{\vartheta_i} I_\mathcal{E}(x) \tilde{f}_{\vartheta_i}(x)dx \\ &\approx \arg\max_\vartheta \hat{\mathbb{E}}_{\tilde{f}_{\vartheta_i}}\left[\log\left(\tilde{f}_\vartheta(x)\right) \tilde{L}_{\vartheta_i}(x) I_\mathcal{E}(x)\right]\end{aligned} \quad (35)$$

where $I_\mathcal{E}(x)$ are samples in the previous iteration and $\hat{\mathbb{E}}_{\tilde{f}_{\vartheta_i}}[\cdot]$ denotes the empirical average.

There are many possible choices for the family of $\tilde{f}_\vartheta(x)$. Here we use a popular class named the Exponential Change of Measure (ECM) for $TTC_L^{-1}$.

Recall that $TTC_L^{-1} \sim \exp\left(\lambda_{TTC_L^{-1}}(v_L)\right)$. ECM considers the family

$$\tilde{f}_{TTC_L^{-1}}(x) = \exp\left(\vartheta_{TTC_L^{-1}}^{ECM} x - \Psi\left(\vartheta_{TTC_L^{-1}}^{ECM}\right)\right) f_{TTC_L^{-1}}(x) \quad (36)$$

parametrized by $\vartheta_{TTC_L^{-1}}^{ECM}$, where $\Psi\left(\vartheta_{TTC_L^{-1}}\right)$ is the logarithmic moment generation function of $TTC_L^{-1}$, i.e.,

$$\Psi\left(\vartheta_{TTC_L^{-1}}^{ECM}\right) = \log \mathbb{E}\left(\exp\left(\vartheta_{TTC_L^{-1}}^{ECM} TTC_L^{-1}\right)\right) \quad (37)$$

It can be derived that

$$\begin{aligned}&\tilde{f}_{TTC_L^{-1}}(x) \\ &= \left(\frac{1}{\lambda_{TTC_L^{-1}}} - \vartheta_{TTC_L^{-1}}^{ECM}\right)\exp\left(-\left(\frac{1}{\lambda_{TTC_L^{-1}}} - \vartheta_{TTC_L^{-1}}^{ECM}\right)x\right)\end{aligned} \quad (38)$$

where $\vartheta_{TTC_L^{-1}}^{ECM} < 1/\lambda_{TTC_L^{-1}}$ and $\lambda_{TTC_L^{-1}} > 0$. To make $\vartheta_{TTC_L^{-1}}^{ECM}$ have the same scale as $\lambda_{TTC_L^{-1}}$, we apply a nonlinear mapping by letting

$$\vartheta_{TTC_L^{-1}}^{ECM} = \frac{\vartheta_{TTC_L^{-1}}}{\vartheta_{TTC_L^{-1}}\lambda_{TTC_L^{-1}} - \lambda_{TTC_L^{-1}}^2} \quad (39)$$

with $\vartheta_{TTC_L^{-1}} < \lambda_{TTC_L^{-1}}$. Substitute Eq. (39) into Eq. (38), we have

$$\begin{aligned}&\tilde{f}_{TTC_L^{-1}}\left(x|\vartheta_{TTC_L^{-1}}\right) \\ &= \left(\frac{1}{\lambda_{TTC_L^{-1}} - \vartheta_{TTC_L^{-1}}}\right)\exp\left(-\frac{x}{\lambda_{TTC_L^{-1}} - \vartheta_{TTC_L^{-1}}}\right)\end{aligned} \quad (40)$$

$R_L^{-1}$ follows a (truncated) Pareto distribution, i.e.

$$f_{R_L^{-1}}(x) = Pareto\left(x\middle|k_{R_L^{-1}}, \sigma_{R_L^{-1}}, \theta_{R_L^{-1}}\right) \quad (41)$$

We apply an ECM of the exponential distribution as our family of IS distributions, where we first construct an exponential distribution

$$\tilde{f}_{R_L^{-1}}(x) = \frac{1}{\lambda_{R_L^{-1}}}\exp\left(-\frac{1}{\lambda_{R_L^{-1}}}x\right) \quad (42)$$

with $\lambda_{R_L^{-1}}$, which gives Eq. (42) the smallest least square error to Eq. (41). With similar procedure, we have

$$\begin{aligned}&\tilde{f}_{R_L^{-1}}\left(x|\vartheta_{R_L^{-1}}\right) \\ &= \left(\frac{1}{\lambda_{R_L^{-1}} - \vartheta_{R_L^{-1}}}\right)\exp\left(-\frac{x}{\lambda_{R_L^{-1}} - \vartheta_{R_L^{-1}}}\right)\end{aligned} \quad (43)$$

Using this approximate ECM instead of an ECM applied to a truncated Pareto reduces the computation complexity in the optimization step since a closed form can be obtained in each Cross Entropy iteration.

The overall likelihood ratio is

$$\begin{aligned}&\tilde{L}(R_L^{-1} = x, TTC_L^{-1} = y) \\ &= \frac{f(x)}{\tilde{f}_{\vartheta_i}(x)} = \frac{f_{R_L^{-1}}(x) f_{TTC_L^{-1}}(y)}{\tilde{f}_{R_L^{-1}}(x) \tilde{f}_{TTC_L^{-1}}(y)}\end{aligned} \quad (44)$$

For low-velocity conditions, i.e.

$$v_L \in (5 \text{ m/s}, 15 \text{ m/s})$$

we simulate N tests with initial condition

$$x = [v_L, TTC_L^{-1}, R_L^{-1}]$$

where $v_L$ follows the low velocity portion (5 m/s~15 m/s) of the empirical distribution shown in Fig. 5. $R_L^{-1}$ and $TTC_L^{-1}$ follows $\tilde{f}_{R_L^{-1}}(x)$ and $\tilde{f}_{TTC_L^{-1}}(x)$. Apply Eqs. (38) and (43) to Eq. (35). The optimal parameter $\vartheta_{R_L^{-1}}$ and $\vartheta_{TTC_L^{-1}}$ can be derived analytically

$$\vartheta_{R_L^{-1}} = \frac{\sum_{j=1}^{N} \tilde{L}(x_j)I_\varepsilon(x_j)(\lambda_{R_L^{-1}} - R_{L,j}^{-1})}{\sum_{j=1}^{N} \tilde{L}(x_j)I_\varepsilon(x_j)} \quad (45)$$

$$\vartheta_{TTC_L^{-1}} = \frac{\sum_{j=1}^{N} \tilde{L}(x_j)I_\varepsilon(x_j)(\lambda_{TTC_L^{-1}} - TTC_L^{-1})}{\sum_{1}^{N} \tilde{L}(x_j)I_\varepsilon(x_j)} \quad (46)$$

where $j$ is the index for each simulation. The newly obtained $\vartheta_{R_L^{-1}}$ and $\vartheta_{TTC_L^{-1}}$ can be used in the next iteration.

The same procedure can be used to obtain optimal parameters in medium and high-velocity conditions.

## IV. SIMULATION ANALYSIS

An AV model was designed to demonstrate the proposed accelerated evaluation approach in the lane change scenarios.

### A. Design of Test Automated Vehicle

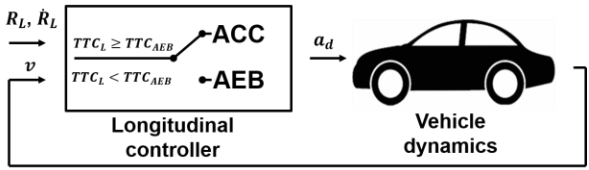

Fig. 11. Layout of the AV model

The AV is designed to be equipped with both Adaptive Cruise Control (ACC) [59] and Autonomous Emergency Braking (AEB). When the driving is perceived to be safe ($TTC_L \geq TTC_{AEB}$), it is controlled by the ACC. The ACC is approximated by a discrete Proportional-Integral (PI) controller [59] to achieve a desired time headway $T_{HW_d}^{ACC}$. Use the time headway error $t_{HW}^{Err}$ as the controller input.

$$t_{HW}^{Err} = t_{HW} - T_{HW_d}^{ACC} \quad (47)$$

where $t_{HW}$ is the current time headway, defined as

$$t_{HW} = R_L/v \quad (48)$$

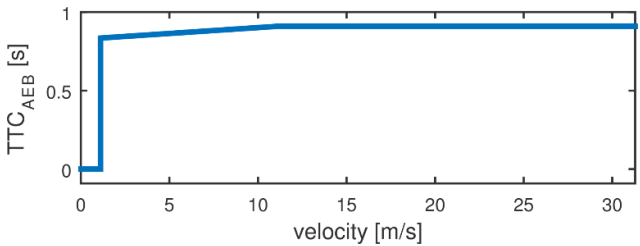

Fig. 12. $TTC_{AEB}$ as a function of vehicle speed

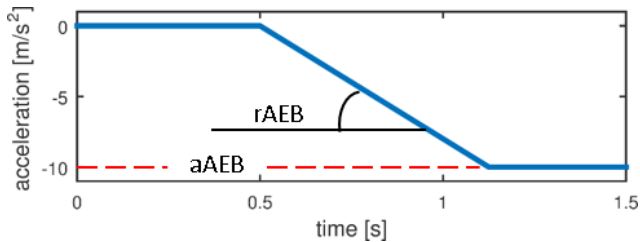

Fig. 13. The modeled AEB algorithm

The discrete PI controller can be described in the discrete-time domain as

$$\frac{A_d(z)}{T_{HW}^{Err}(z)} = K_p^{ACC} + K_i^{ACC}\frac{T_s}{2}\frac{z+1}{z-1} \quad (49)$$

where $A_d$ and $T_{HW}^{Err}$ are the Z transformation of the command acceleration $a_d$ and $t_{HW}^{Err}$; $T_s$ is the sampling time; gains $K_p^{ACC}$ and $K_i^{ACC}$ are calculated using the Matlab Control Toolbox using the following requirements: a) Loop bandwidth = 10 rad/s, and b) Phase margin = 60 degree. The control power of ACC system is saturated to a constant acceleration $a_{ACC}^{Max}$, i.e. $|a_d| \leq a_{ACC}^{Max}$. To implement the PI controller in the time domain, taking the inverse Z transformation of Eq. (49), we get

$$a_d(k+1) = a_d(k) + K_p^{ACC}\big(t_{HW}^{Err}(k) - t_{HW}^{Err}(k-1)\big) \\ + K_i^{ACC}\big(t_{HW}^{Err}(k) + t_{HW}^{Err}(k-1)\big)T_s/2 \quad (50)$$

The AEB model was extracted from a 2011 Volvo V60, based on a test conducted by ADAC (Allgemeiner Deutscher Automobil-Club e.V.) [60]. It is analyzed using test track data, owner's manuals, European New Car Assessment Program (Euro NCAP) information, and videos during vehicle operation [61]. The AEB algorithm becomes active when $TTC_L < TTC_{AEB}$, where $TTC_{AEB}$ depends on the vehicle speed as shown in Fig. 12. Once triggered, AEB aims to achieve acceleration $a_{AEB}$. In [61] $a_{AEB}$ was assumed to be -10 m/s² on high friction roads. The build-up of deceleration is subject to a rate limit $r_{AEB}$ as shown in Fig. 13. It should be noted that the AEB modeled here is an approximation but not necessarily a good representation of the actual AEB system on production vehicles.

A first order lag with a time constant $\tau_{AV}$ is used to model the transfer function from the commanded acceleration to the actual acceleration for simplicity. The proposed accelerated evaluation process can be applied on other vehicle models such as CarSim [62], if more accurate simulations are desired.

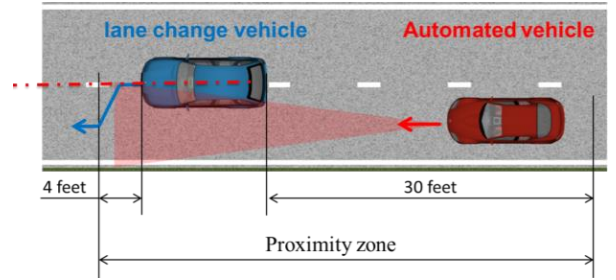

Fig. 14. Definition of conflict events

The simulation parameters are listed in TABLE I.

TABLE I
PARAMETERS FOR THE LANE CHANGE SIMULATIONS

| Var. | $T_{HW_d}^{ACC}$ | $a_{ACC}^{Max}$ | $K_p^{ACC}$ | $K_i^{ACC}$ | $a_{AEB}$ | $r_{AEB}$ | $\tau_{AV}$ | $T_s$ | $T_{LC}$ |
|---|---|---|---|---|---|---|---|---|---|
| Unit | s | m/s² | - | - | m/s² | m/s³ | s | s | s |
| Value | 2 | 5 | −38.6 | −1.35 | 10 | -16 | 0.0796 | 0.1 | 8 |

### B. Simulation Analysis

Three kinds of events were analyzed in this study:
- Conflict
- Crash

- Injury

A conflict event happens when an AV appears in the proximity zone of the LCV between time $t_{LC}$ and $t_{LC} + T_{LC}$. As shown in Fig. 14, the proximity zone is the area in the adjacent lane from 4 feet in front of the bumper of the LCV to 30 feet behind the rear bumper of the LCV [42, p. ix]. This area generally includes the blind spot and the area beside and behind the vehicle in which another vehicle is likely to travel.

The Cross Entropy is used to find optimal $\vartheta_{TTC_L^{-1}}$ and $\vartheta_{R_L^{-1}}$. The values of $\vartheta_{TTC_L^{-1}}$ and $\vartheta_{R_L^{-1}}$ in the tenth iteration are used in the simulations to estimate the probability of conflicts (conflict rate) in a lane change scenario. 100 lane changes are simulated in each iteration. As shown in Fig. 15, three sets of $\vartheta_{TTC_L^{-1}}$ and $\vartheta_{R_L^{-1}}$ are obtained with low, medium and high velocities. All $\vartheta_{R_L^{-1}}$ converge to about -0.12, whereas values of $\vartheta_{TTC_L^{-1}}$ float around zero. As the conflict events are defined based on $R_L$, $R_L$ has a direct impact on the occurrence of the event. Therefore $\tilde{f}_\vartheta(x)$ is largely affected by $f_{R_L^{-1}}(x)$, and $\vartheta_{TTC_L^{-1}}$ converges to zero $\left(\tilde{f}_{TTC_L^{-1}}(x) \approx f_{TTC_L^{-1}}(x)\right)$.

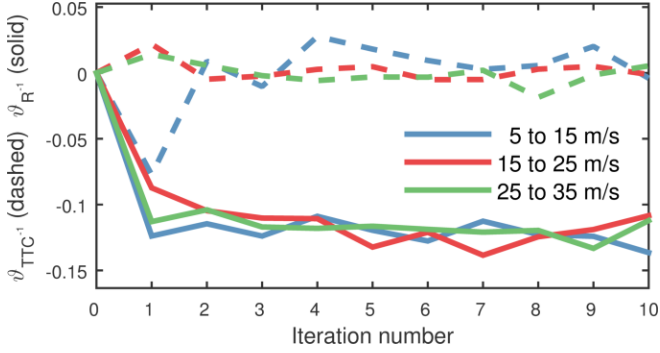

Fig. 15. Searching for optimal parameters for conflict events

Both accelerated evaluation and the non-accelerated simulations (based on CMC) were conducted to demonstrate the performance and credibility of the proposed approach. Fig. 16 shows that the accelerated test is unbiased as the conflict rate converges to the one estimated in the non-accelerated simulation.

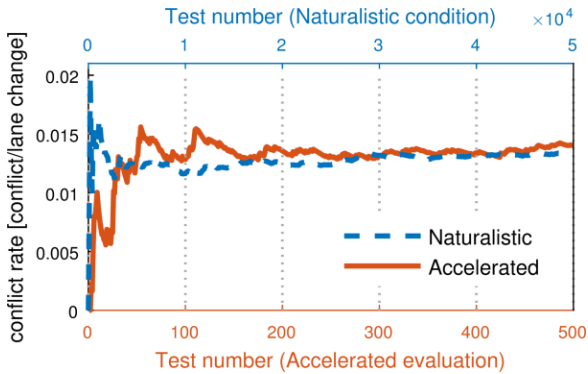

Fig. 16. Estimation of the conflict rate

The convergence is reached when the relative half-width $l_r$ is below $\beta = 0.2$ with 80% confidence. Fig. 17 shows that the accelerated evaluation achieves this confidence level after $N_{acc} = 364$ simulations, while the naturalistic simulations take $N_{nature} = 5.90e3$ simulations.

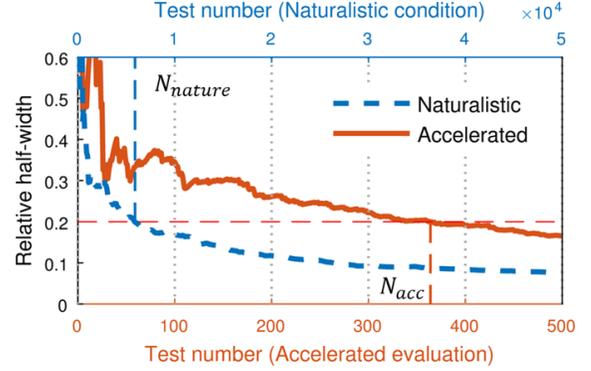

Fig. 17. Convergence of conflict rate estimation

In the SPMD database, during 1,325,964 miles naturalistic driving, 173,592 lane changes were identified with negative range rates. The frequency of lane change can be estimated as

$$r_{lc} = \frac{1,325,964}{173,592} = 7.64 \; [\text{mile/lane change}] \quad (51)$$

The driving distance needed in naturalistic test is thus

$$D_{nature} = r_{lc} \cdot N_{nature} \quad (52)$$

The test distance in accelerated evaluation

$$D_{acc} = \sum_{n=1}^{N_{acc}} \int_{t=t_{LC}}^{t_{LC}+\mathcal{T}_{LC}} v^{(n)}(t) \, dt \quad (53)$$

where $v^{(n)}(t)$ represents the velocity of AV at time t in the $n_{th}$ test and the termination time

$$\mathcal{T}_{LC} = \min\{min(t|R_L(t) < R_\mathcal{E}), T_{LC}\} \quad (54)$$

The accelerated rate is defined as

$$r_{acc} = \frac{D_{nature}}{D_{acc}} \quad (55)$$

The acceleration is achieved from both the modeling of lane change scenarios and the application of Importance Sampling and Cross Entropy techniques.

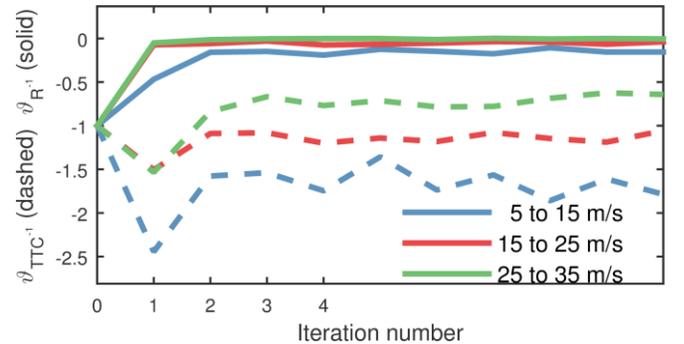

Fig. 18. Searching for optimal parameters for crash events

A crash happens when the range $R_L$ becomes negative, i.e. $R_\mathcal{E} = 0$ in Eq. (24). Similar to the conflict events analysis, another Cross Entropy analysis is conducted to find optimal $\vartheta_{TTC_L^{-1}}$ and $\vartheta_{R_L^{-1}}$ for crash events. Because crashes are rarer than the conflict events, 500 lane changes are simulated in each iteration. As shown in Fig. 18, three different values of $\vartheta_{TTC_L^{-1}}$ were obtained from the iterative search for different velocity intervals, whereas $\vartheta_{R_L^{-1}}$ converges to values close to zero. It can be explained that in the crash analysis, the safety critical function (AEB) on AV is mainly affected by TTC. Therefore $\vartheta_{TTC_L^{-1}}$ has a larger impact on the occurrence of the crash. The estimation of the crash rate under accelerated and naturalistic conditions are shown in Fig. 19. The convergence is reached with 80% confidence level and $\beta = 0.2$ as shown in Fig. 20.

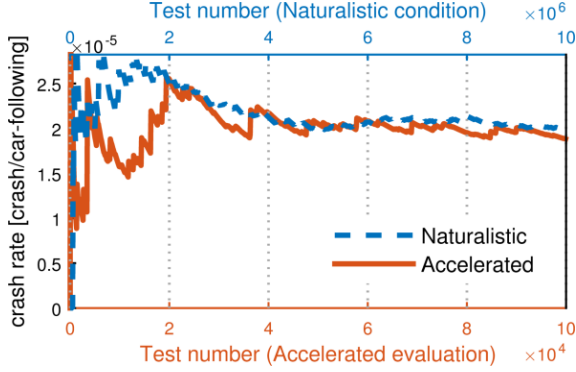

Fig. 19. Estimation of the crash rate

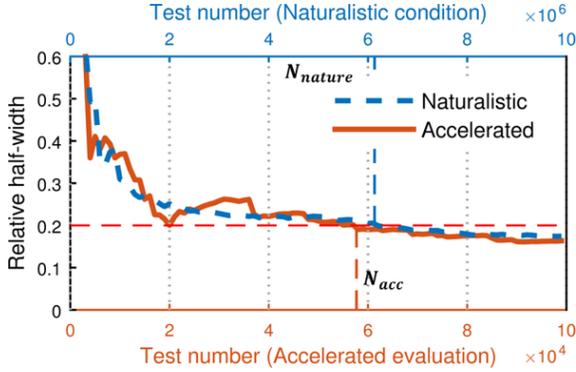

Fig. 20. Convergence of crash rate estimation

Injuries are also important indicators of the performance of AVs. Here we focus on injuries with the Maximum Abbreviated Injury Score equal or larger than 2 (MAIS2+), representing moderate-to-fatal injuries. The probability of injury is related to the relative velocity at the crash time $t_{crash}$

$$\Delta v = -\dot{R}_L(t_{crash}) > 0 \quad (56)$$

The probability of moderate-to-fatal injuries for the AV passengers is estimated by a nonlinear model

$$P_{inj}(\Delta v) = \begin{cases} \dfrac{1}{1 + e^{-(\beta_0 + \beta_1 \Delta v + \beta_2)}} & crash \\ 0 & no\ crash \end{cases} \quad (57)$$

which was proposed by Kusano and Gabler [63] shown in Fig. 21 with parameters $\beta_0 = -6.068$, $\beta_1 = 0.1$, and $\beta_2 = -0.6234$. The injury rate $\mathrm{E}\left(P_{inj}(\Delta v)\right)$ is calculated as

$$\mathrm{E}\left(P_{inj}(\Delta v)\right) = \hat{\mathrm{E}}_{f^*}\left(P_{inj}(\Delta v)\right) \quad (58)$$

$$\approx \frac{1}{n}\sum_{i=0}^{N_{acc}} P_{inj}(\Delta v(x_n)) L(x_n)$$

where $L$ is the likelihood and $x_n$ represents the random variables ($[v_L, TTC_L^{-1}, R_L^{-1}]$) in the $n^{th}$ simulation. The modified statistics used in crash events (shown in Fig. 18) are used to calculate the injury rate. The estimation results and convergence are shown in Fig. 22 and Fig. 23.

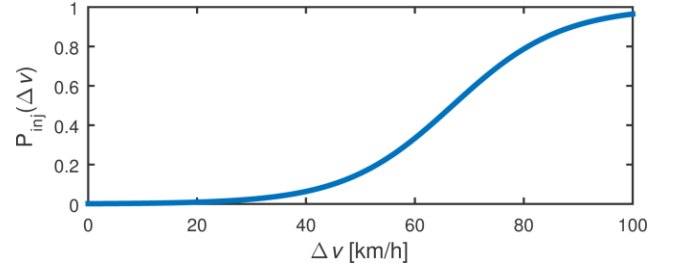

Fig. 21. Moderate-to-fatal injury model for forward collisions

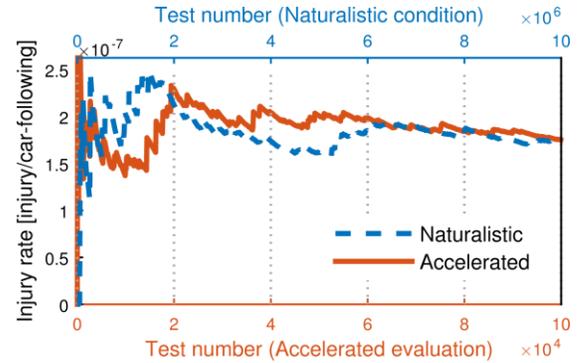

Fig. 22. Estimation of injury rate

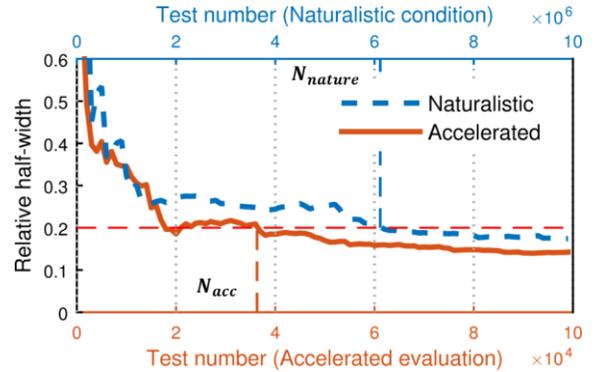

Fig. 23. Convergence of injury rate estimation

The accelerated rates of conflict, crash and injury events are summarized in TABLE II. The accelerated rates of crashes and injuries are higher than that of conflicts. This is because crashes and injuries occur with much lower probabilities than conflicts. The IS techniques generally have better performance when target events are rarer.

TABLE II
ACCELERATED RATES OF CONFLICTS, CRASHES AND INJURY

|  | Conflict | Crash | Injury |
| --- | --- | --- | --- |
| $D_{nature}$ [mile] | 4.53e4 | 4.71e7 | 4.70e7 |
| $D_{acc}$ [mile] | 16.4 | 4.02e3 | 2.53e3 |
| $r_{acc}$ | 2.77e3 | 1.17e4 | 1.86e4 |

## V. Conclusion

This paper proposes a new approach to evaluate the performance of AVs in an accelerated fashion. A lane change model was established based on a large naturalistic driving database – the Safety Pilot Model Deployment database. Lane change conflict, crash, and injury rates of a given AV model were estimated accurately but 2,000 to 20,000 times faster than the naturalistic driving tests in simulation. This technique thus has the potential to reduce greatly the development and validation time for AVs by providing both statistical conclusion and critical scenarios selected objectively.

In the future study, more comprehensive human-controlled model may be obtained as more data are collect in the Safety Pilot Model Deploy project and other projects. Other forms of IS distribution families other than ECM-based will be analyzed to potentially increase the evaluation efficiency to an even higher rate. The proposed accelerated evaluation approach can also be extended to other scenarios, such as car-following, lane departure or pedestrian avoidance and other testing platforms in addition to pure simulations, such that hardware-in-the-loop tests, driving simulator tests, or on-track tests.

## Disclaimers

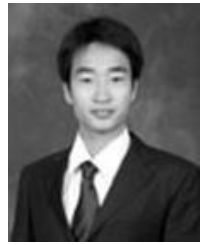

**Ding Zhao** received the Ph.D. degree in 2016 from the University of Michigan, Ann Arbor. He is currently a Research Fellow in the University of Michigan Transportation Research Institute. His research interest includes automated vehicles, connected vehicles, driver modeling, and big data analysis.

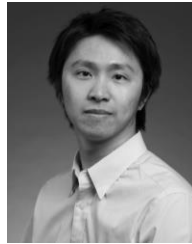

**Henry Lam** received the B.S. degree in actuarial science from the University of Hong Kong in 2005, and the A.M. and Ph.D. degrees in statistics from Harvard University, Cambridge, in 2006 and 2011.

From 2011 to 2014, he was an Assistant Professor in the Department of Mathematics and Statistics at Boston University. Since 2015, he has been an Assistant Professor in the Department of Industrial and Operations Engineering at the University of Michigan, Ann Arbor. His research focuses on stochastic simulation, risk analysis, and simulation optimization. Dr. Lam's works have been funded by National Science Foundation and National Security Agency. He has also received an Honorable Mention Prize in the Institute for Operations Research and Management Sciences (INFORMS) George Nicholson Best Student Paper Award, and Finalist in INFORMS Junior Faculty Interest Group Best Paper Competition.

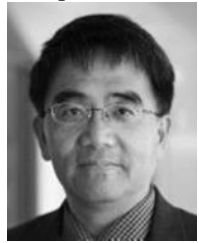

**Huei Peng** received the Ph.D. degree from the University of California, Berkeley, CA, USA, in 1992. He is currently a Professor with the Department of Mechanical Engineering, University of Michigan, Ann Arbor, MI, USA. He is currently the U.S. Director of the Clean Energy Research Center—Clean Vehicle Consortium, which supports 29 research projects related to the development and analysis of clean vehicles in the U.S. and in China. He also leads an education project funded by the Department of Energy to develop ten undergraduate and graduate courses, including three laboratory courses focusing on transportation electrification. He has more than 200 technical publications, including 85 in refereed journals and transactions. His research interests include adaptive control and optimal control, with emphasis on their applications to vehicular and transportation systems. His current research focuses include design and control of hybrid vehicles and vehicle active safety systems.

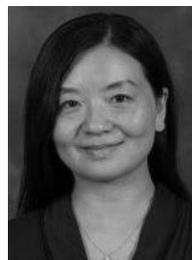

**Shan Bao** received the B.E. and M.E. degree in mechanical engineering from Hefei University of Technology, China and Ph.D. degrees in Industrial Engineering from University of Iowa, Iowa City, IA.

Dr. Bao is currently an assistant research scientist in UMTRI's Human Factors Group. She joined UMTRI in 2009, starting as a postdoctoral fellow after completing her Ph.D. in industrial engineering at the University of Iowa. Her research interests focus on driver behavior modeling, driver distraction, naturalistic driving data analysis and driver-simulator study.

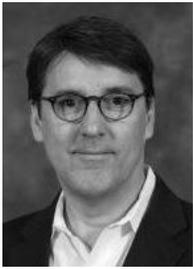
**Dave LeBlanc** received a Ph.D. in aerospace engineering from the University of Michigan, and master's and bachelor's degrees in mechanical engineering from Purdue University.

Dr. David J. LeBlanc is currently an associate research scientist, has been at UMTRI since 1999. Dr. LeBlanc's work focuses on the automatic and human control of motor vehicles, particularly the design and evaluation of driver assistance systems.

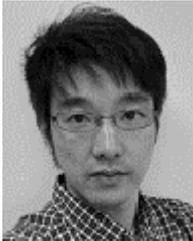
**Kazutoshi Nobukawa** received the B.E. and M.E. degrees in materials science and engineering from Waseda University, Tokyo, Japan, and M.S.E. and Ph.D. degrees in mechanical engineering from University of Michigan, Ann Arbor, MI.

From 2008 to 2010, he was a Graduate Student Research Assistant with the University of Michigan Transportation Research Institute (UMTRI), Ann Arbor, MI. From 2012, He has been a Research Fellow in the UMTRI's Engineering Systems Group. His research interest includes modeling and control of dynamical systems for analysis of collision avoidance systems, vehicle dynamics, tracking, and data mining.

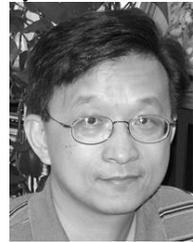
**Christopher S. Pan** is a senior researcher with the National Institute for Occupational Safety and Health (NIOSH), in Morgantown, West Virginia. He received his M.S. in 1989 and Ph.D. in 1991 in Industrial Engineering from the University of Cincinnati and has been conducting research at NIOSH since 1989, with projects chiefly focusing on ergonomics/safety. He currently has an appointment as an adjunct professor in the Department of Industrial and Management Systems Engineering at West Virginia University. He currently serves as a project officer at NIOSH for six funded studies in construction and transportation sectors, including a follow-up collaborative project (2013–2015) of a motor vehicle study with the University of Michigan Transportation Research Institute (UMTRI). For these and related research endeavors, he has been recognized by distinguished peers and professionals in the occupational safety and health community as a competent safety professional, project manager, ergonomist, inventor, and scientist.